%% file: main.tex
\documentclass[10pt,twocolumn,letterpaper]{article}

\usepackage[pagenumbers]{cvpr} %

\input{preamble}

\definecolor{cvprblue}{rgb}{0.21,0.49,0.74}
\usepackage[pagebackref,breaklinks,colorlinks,allcolors=cvprblue]{hyperref}

\def\confName{CVPR}
\def\confYear{2027}

\title{\methodname: A Synthetic Indoor World Dataset}

\input{authors}

\begin{document}
\twocolumn[{%
\renewcommand\twocolumn[1][]{#1}%
\maketitle
\input{figures/teaser}

}]

\begingroup
  \renewcommand{\thefootnote}{\fnsymbol{footnote}}
  \footnotetext[1]{Equal contribution.}
\endgroup

\input{sec/0_abstract}
\input{sec/1_intro_anpei}
\input{sec/2_related}
\input{sec/3_method_anpei}
\input{sec/4_dataset_anpei}
\input{sec/5_experiments_anpei}
\input{sec/6_conclusion}
\input{sec/7_ack}
\clearpage
\clearpage
{
    \small
    \bibliographystyle{ieeenat_fullname}
    \bibliography{habi}
}

\end{document}

%% file: preamble.tex
\pretocmd{\abstract}{\setlength{\parskip}{0pt}}{}{\PackageError{habi}{Cannot adjust Abstract paragraph spacing}{Check the CVPR abstract definition.}}

\usepackage{needspace}
\usepackage{booktabs}
\usepackage{tabularx}
\usepackage{stfloats}
\usepackage{multirow}
\usepackage{pifont}
\usepackage{tikz}
\usetikzlibrary{arrows.meta,positioning,fit,calc,shapes.misc}

\definecolor{memteal}{HTML}{008D81}
\definecolor{viewblue}{HTML}{3468B2}
\definecolor{mutedink}{HTML}{566174}
\definecolor{habires}{HTML}{3468B2}
\definecolor{habinonres}{HTML}{E08A2E}
\definecolor{habihouse}{HTML}{7A5BB5}
\definecolor{habiartist}{HTML}{2E9E6A}
\definecolor{habiconv}{HTML}{8A8F98}

\newcommand{\methodname}{HABI-12K\xspace}
\newcommand{\bench}{\textup{\textsc{HABI-Bench}}\xspace}
\newcommand{\cmark}{\ding{51}}
\newcommand{\xmark}{\ding{55}}
\newcommand{\pmark}{$\circ$}

\xspaceaddexceptions{\,}
\input{stats}

\usepackage{microtype}

\newcommand{\figref}[1]{Fig.~\ref{#1}}

\newcommand{\tabref}[1]{Tab.~\ref{#1}}

%% file: stats.tex
\newcommand{\NumScenes}{12{,}081\xspace}           %
\newcommand{\NumFrames}{2.16M\xspace}              %
\newcommand{\NumFramesExact}{2{,}158{,}764\xspace}
\newcommand{\NumHours}{300\xspace}                 %
\newcommand{\NumKm}{838\xspace}                    %

\newcommand{\MedFrames}{113\xspace}

\newcommand{\MedDuration}{56\xspace}
\newcommand{\MedPath}{45.1\xspace}

\newcommand{\PDepthTwo}{54\%\xspace}
\newcommand{\PDepthFive}{92\%\xspace}

\newcommand{\MedLum}{0.45\xspace}

\newcommand{\NumMajor}{15\xspace}
\newcommand{\NumMinor}{75\xspace}

\newcommand{\GeoMedRel}{0.06\%\xspace}

%% file: authors.tex
\definecolor{projectlink}{cmyk}{0,1,0,0}
\author{%
  \large
  Sibo Wu\textsuperscript{*}\quad
  Xiaoqian Liang\textsuperscript{*}\quad
  Zidu Wang\quad
  Anpei Chen\\[8pt]
  \normalsize
  Inception3D Lab, Westlake University
}

%% file: figures/teaser.tex
\vspace{-1.2em }
\begingroup\centering
\captionsetup{type=figure,hypcap=false}
\includegraphics[width=\textwidth]{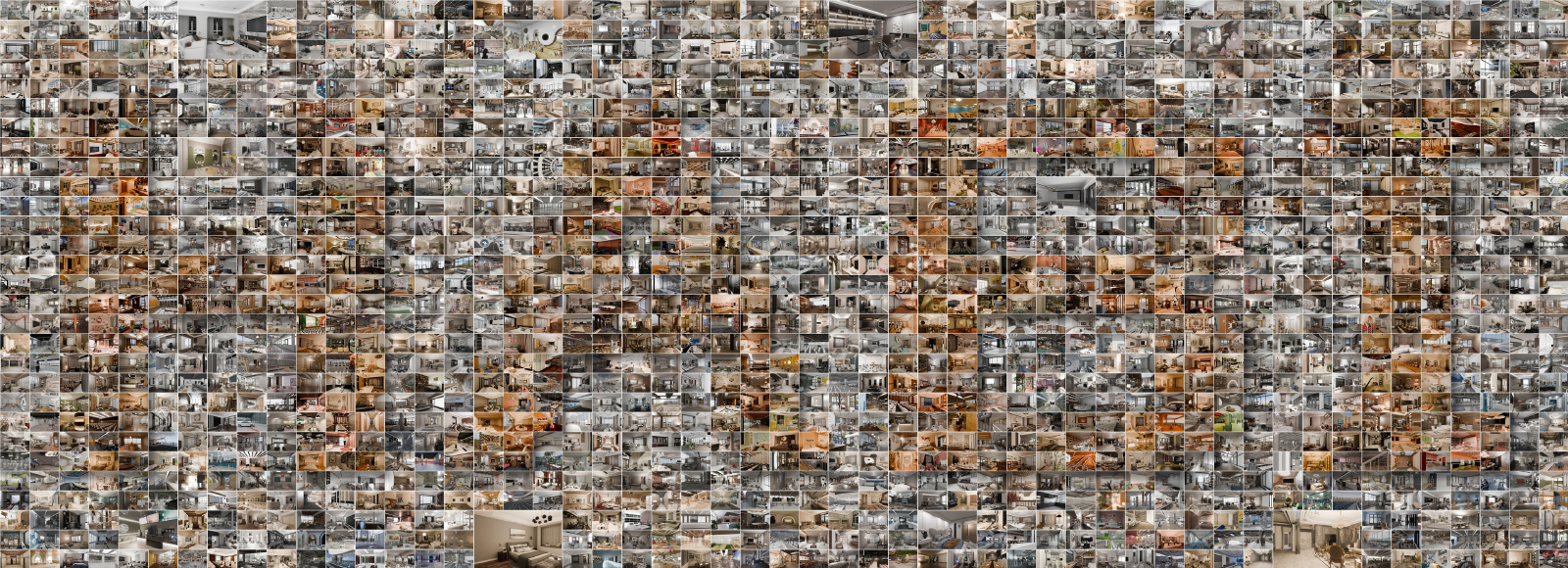}
\vspace{-1.2em}
\captionof{figure}{We introduce \textbf{\methodname}, a large-scale synthetic dataset spanning diverse residential, commercial, and public scenes.
It contains \NumScenes scenes across \NumMajor major and \NumMinor minor categories, with 2.16 million $2048\times1024$ path-traced panoramas rendered along continuous camera trajectories.
Aligned RGB, metric depth, shading normals, and exact camera poses support benchmarks for multi-view geometry and novel view synthesis, as well as learning 3D scene representations at scale.}
\label{fig:teaser}
\par\endgroup
\vspace{0.6em}

%% file: sec/0_abstract.tex
\begin{abstract}
\looseness=-1
We introduce HABI-12K, a large-scale synthetic dataset for scene-level 3D learning, comprising 12,081 indoor scenes across 15 major categories and 75 minor categories.
The collection covers residential, commercial, and public spaces drawn from professionally authored interiors.
To consolidate these environments, we convert scenes from heterogeneous sources, including 3ds Max and Unity, into a common Blender format, enabling consistent scene processing, rendering, and geometric annotation.
An agent-in-the-loop pipeline supports scene-specific repairs, camera trajectory planning, and deterministic quality checks.
Along continuous camera trajectories, we render 2.16 million path-traced panoramas paired with metric ray depth, world-space shading normals, and exact camera poses.
Full-sphere observations allow perspective views with different orientations and fields of view to be sampled at the same camera positions without additional rendering, supporting flexible input configurations across tasks.
We further introduce \bench, which evaluates geometry models and view synthesis models on 117 test scenes, measuring camera estimation, depth prediction, and novel view synthesis across scene groups and input-view budgets. The dataset is available in   {\hypersetup{urlcolor=projectlink}\small\usefont{T1}{pcr}{m}{n}\urlstyle{same}\url{https://huggingface.co/datasets/Inception3D/HABI-12K}}.%

\end{abstract}

%% file: sec/1_intro_anpei.tex
\input{tables/dataset_comparison}

\section{Introduction}
\label{sec:habi_introduction}

Learning indoor reconstruction and view synthesis requires realistic images paired with consistent geometry across viewpoints.
Depth estimation, reconstruction, and view synthesis require different view counts, directions, and degrees of overlap.
Supporting these settings calls for data spanning diverse scene categories and spatial configurations.

\looseness=-1
Existing datasets provide complementary foundations.
Real-world collections such as DL3DV-10K~\cite{data_dl3dv} and ScanNet++~\cite{yeshwanth2023scannetpp} capture scenes, with supervision determined by capture and reconstruction.
Hypersim~\cite{data_hypersim}, Structured3D~\cite{zheng2020structured3d}, and InteriorNet~\cite{li2018interiornet} provide scene-derived supervision by rendering professionally authored interiors.
Yet none of them combines dense depth and normals, continuous trajectories, panoramas, and non-residential venues (\tabref{tab:dataset-comparison}): the real captures lack dense depth or normals, while the synthetic ones are limited to a few hundred scenes, isolated viewpoints, or residential layouts.
Building on these resources, we aim to capture diverse residential, commercial, and public spaces through realistic, geometry-supervised camera trajectories.

\looseness=-1
We introduce \textbf{HABI-12K}, comprising 12,081 scenes and 2.16 million path-traced panoramas, primarily drawn from professionally authored interiors (\figref{fig:teaser}).
Twelve commercial and public-space semantic categories account for 6,907 scenes (57.2\%), including hotels, restaurants, offices, shops, gyms, and auditoriums.
\figref{fig:categories} shows representative perspective views across all \NumMajor major scene categories.
Along continuous camera trajectories, each panorama pairs RGB with metric ray depth, world-space shading normals, and scene-derived camera parameters.
At each sampled camera position, full-sphere observations support perspective views with different orientations and fields of view without additional rendering.

Professional interiors supply detailed architecture, furnishings, and coordinated materials.
Assets authored for selected presentation views, however, may contain incomplete geometry, disconnected rooms, or lighting unsuitable for moving cameras.
Using these assets along trajectories therefore requires scene-specific preparation and validation.
Our agent-in-the-loop pipeline converts heterogeneous assets, assembles multi-room houses, plans camera trajectories, and repairs lighting (\figref{fig:pipeline}).
Agents diagnose failures from geometric reports and previews; deterministic programs execute and verify their proposed repairs, followed by a frame-by-frame RGB and depth audit.

\input{figures/category_gallery}

We introduce \textbf{\bench}, evaluating nine geometry and nine feed-forward view synthesis models on 117 evaluation scenes across view budgets and scene groups.
View selection uses ground-truth visibility, and all methods within each task use the same input lists and evaluation views.
Metrics cover camera and depth accuracy and novel view rendering quality.
Additional context improves depth on fixed anchor views for all nine geometry models, although full-set error increases for most models as more views are evaluated.
Fixed-view evaluation thus separates context benefits from changes in evaluated content.

In summary, we make the following contributions:
\begin{itemize}
    \item \looseness=-1 A dataset of 12,081 scenes covering residential, commercial, and public spaces, with 2.16 million geometry-supervised panoramas supporting flexible perspective view sampling.
    \item An agent-in-the-loop production pipeline that adapts heterogeneous design assets for observations along camera trajectories through scene repair, trajectory planning, and explicit quality checks.
    \item \bench, with protocols and baseline results for camera estimation, depth prediction, and novel view synthesis across residential and non-residential scenes.
\end{itemize}

%% file: tables/dataset_comparison.tex
\begin{table*}[t]
\centering
\caption{\textbf{Comparison with 3D datasets.}
``\# Frames'': released images; ``--'': only meshes, layouts, or Gaussian splats are released.
``Trajectory'': continuous camera motion with poses; ``Multi-room'': scenes spanning connected rooms; ``Non-residential'': commercial or public venues (``--'' indicates unspecified coverage).
\pmark: partially available (\eg derived from meshes, cube maps, or only for a subset of scenes).
$^\ast$Frames with laser-scanner depth. InteriorNet counts are given separately for trajectory sequences and independently sampled layouts.
\methodname frames are $2048\times1024$ panoramas, from which perspective views of any orientation and field of view can be resampled.}
\label{tab:dataset-comparison}
\footnotesize
\setlength{\tabcolsep}{4.2pt}
\resizebox{\textwidth}{!}{%
\begin{tabular}{@{}l l r r c c c c c c@{}}
\toprule
Dataset & Type & \# Scenes & \# Frames & Depth & Normal & Trajectory & Panorama & Multi-room & Non-residential \\
\midrule
Stanford 2D-3D-S~\cite{armeni2017joint} & real & 270 rooms & 72K & \cmark & \cmark & \xmark & \cmark & \cmark & \cmark \\
Matterport3D~\cite{chang2017matterport3d} & real & 90 (2,056 rooms) & 194K & \cmark & \pmark & \xmark & \pmark & \cmark & \pmark \\
ScanNet~\cite{data_scannet} & real & 1,513 & 2.5M & \cmark & \xmark & \cmark & \xmark & \pmark & \cmark \\
ARKitScenes~\cite{data_arkitscenes} & real & 1,661 & 450K$^\ast$ & \cmark & \xmark & \cmark & \xmark & \xmark & \xmark \\
ScanNet++~\cite{yeshwanth2023scannetpp} & real & 460 & 3.98M & \cmark & \xmark & \cmark & \xmark & \pmark & \cmark \\
DL3DV-10K~\cite{data_dl3dv} & real & 10,510 & 51.2M & \xmark & \xmark & \cmark & \xmark & \pmark & \cmark \\
\midrule
Replica~\cite{straub2019replica} & scan, rendered & 18 & -- & \pmark & \pmark & \xmark & \xmark & \pmark & \pmark \\
InteriorNet~\cite{li2018interiornet} & synthetic & 10K / 1.7M & 15M / 5M & \cmark & \pmark & \cmark & \pmark & \cmark & \xmark \\
Structured3D~\cite{zheng2020structured3d} & synthetic & 3,500 (21,835 rooms) & 196.5K & \cmark & \cmark & \xmark & \cmark & \cmark & \xmark \\
3D-FRONT~\cite{fu20213dfront} & synthetic & 6813 & -- & \pmark & \pmark & \xmark & \xmark & \cmark & -- \\
Hypersim~\cite{data_hypersim} & synthetic & 461 & 74.6K & \cmark & \cmark & \cmark & \xmark & \xmark & \cmark \\
HSSD-200~\cite{khanna2024hssd} & synthetic & 211 & -- & \xmark & \xmark & \xmark & \xmark & \cmark & \xmark \\
ASE~\cite{data_ase} & synthetic & 100K & 58M & \cmark & \xmark & \cmark & \xmark & \cmark & \xmark \\
MegaSynth~\cite{jiang2025megasynth} & procedural & 700K & 48 / scene & \cmark & \xmark & \xmark & \xmark & \xmark & -- \\
InternScenes~\cite{zhong2025internscenes} & mixed & 39,870 & -- & \xmark & \xmark & \xmark & \xmark & \pmark & \xmark \\
InteriorGS~\cite{miao2026interiorgs} & synthetic & 1,000 & -- & \xmark & \xmark & \xmark & \xmark & \cmark & \cmark \\
Realsee3D~\cite{li2026argus} & synth.\ + real & 10,000 & 299K & \cmark & \xmark & \xmark & \cmark & \cmark & \xmark \\
\midrule
\textbf{\methodname (ours)} & synthetic & \NumScenes & \NumFrames & \cmark & \cmark & \cmark & \cmark & \cmark & \cmark \\
\bottomrule
\end{tabular}}

\vspace{-0.5em}
\end{table*}

%% file: figures/category_gallery.tex
\begin{figure*}[!t]
\centering
\includegraphics[width=\textwidth]{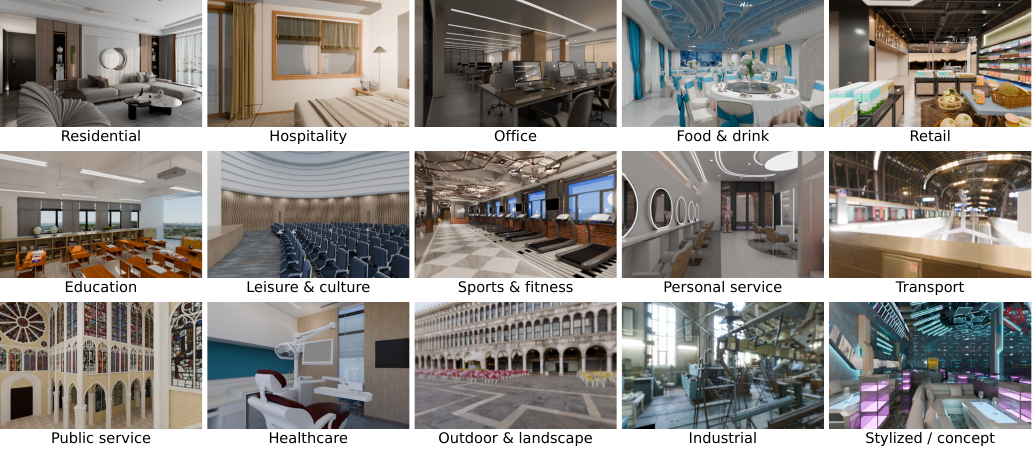}
\vspace{-1.6em}
\caption{\textbf{Visual diversity of \methodname.} Representative perspective views of all \NumMajor major categories.}
\label{fig:categories}
\vspace{-0.6em}
\end{figure*}

%% file: sec/2_related.tex
\section{Related Work}
\label{sec:related}

\paragraph{Real-world Scene Datasets.}
Real-world datasets preserve the appearance and spatial complexity of physical environments, while their geometric supervision depends on the acquisition process.
Video collections such as DL3DV-10K~\cite{data_dl3dv} supply diverse multi-view observations at scale; RGB-D and laser-assisted collections such as ScanNet~\cite{data_scannet}, ScanNet++~\cite{yeshwanth2023scannetpp}, and ARKitScenes~\cite{data_arkitscenes} pair images with measured geometry.
Panoramic capture broadens angular coverage, from discrete scanning positions in Matterport3D~\cite{chang2017matterport3d} to continuous trajectories in Holo360D~\cite{ou2026holo360d}.
Reconstructed meshes in Gibson~\cite{xia2018gibson}, HM3D~\cite{ramakrishnan2021hm3d}, and Replica~\cite{straub2019replica} further permit novel viewpoints to be rendered, although these observations inherit the coverage and completeness of the reconstruction.
Across these routes, supervision depends on what sensors observe and how those observations are registered or reconstructed.
Synthetic scenes provide a complementary route: camera parameters and geometric annotations can be obtained directly from the scene used to render each image.

\paragraph{Procedural and Generative Scene Construction.}
Scaling scene construction requires diversity in both spatial arrangements and the assets that populate them.
Procedural approaches such as SceneNet RGB-D~\cite{mccormac2017scenenet}, ProcTHOR~\cite{deitke2022procthor}, and Infinigen Indoors~\cite{raistrick2024infinigenindoors} vary layouts and objects through rules and sampling, while learned and language-guided methods such as ATISS~\cite{paschalidou2021atiss}, Holodeck~\cite{yang2024holodeck}, and PhyScene~\cite{yang2024physcene} condition arrangements on floor plans, descriptions, or physical constraints.
Recombining a restricted asset vocabulary can yield many layouts while repeating object shapes and styles; procedural assets likewise vary within the design space of their generators.
Agentic systems such as SAGE~\cite{xia2026sage} and SceneSmith~\cite{pfaff2026scenesmith} expand this space through asset synthesis and iterative scene refinement, although object geometry and appearance still depend on the underlying generators or retrieved assets.
Thus, scaling the number of generated scenes does not by itself establish diversity in furnishings, fine geometric detail, or material composition.
\methodname primarily draws on professionally authored interiors, retaining their architectural details, furnishings, and coordinated materials as existing scene content.
This provides a source of variation in both individual assets and their designed combinations across residential, commercial, and public interiors.
Our agents adapt these heterogeneous scenes for moving cameras through diagnosis and repair, with deterministic validation of the resulting scenes and observations.

\paragraph{Photorealistic Synthetic Indoor Datasets.}
PBRS~\cite{zhang2017pbrs}, Hypersim~\cite{data_hypersim}, and InteriorVerse~\cite{zhu2022learning} use physically based rendering to generate indoor images with aligned supervision. OpenRooms~\cite{li2021openrooms} reconstructs layouts from indoor scans and renders them with aligned CAD furniture and surface materials.
Within this setting, scaling the scene collection and broadening its semantic coverage are separate goals.
Large collections of residential layouts in InteriorNet~\cite{li2018interiornet} and Structured3D~\cite{zheng2020structured3d} support variation within homes; Realsee3D~\cite{li2026argus} similarly emphasizes residential environments, combining 9K synthetic and 1K real scenes with 299K panoramic viewpoints.
Commercial and public interiors are represented in Hypersim~\cite{data_hypersim} and InteriorGS~\cite{miao2026interiorgs}, but across smaller collections of 461 and 1,000 scenes, respectively.
Observation count is a further, distinct dimension: rendering more views expands coverage within an environment without increasing the number of independent environments.
This distinction matters for panoramic resources: Matrix-Pano~\cite{yang2025matrix3d} provides 116,759 video sequences from 504 Unreal Engine environments, whereas PanoX~\cite{huang2026omnix} supplies geometry and material annotations from eight environments.
Other panoramic datasets expand their sources by combining scans and synthetic assets, as in CM-EVS~\cite{liu2026cmevs}, or through procedural generation, as in PanoInfinigen~\cite{bozic2026pager} and the panoramas used by Stepper~\cite{wimbauer2026stepper}; CM-EVS curates 36,373 views from 1,275 indoor scenes, while PanoInfinigen spans 20,000 indoor and natural scenes.
These resources show that scene scale, appearance, semantic coverage, and observation density must be assessed together.
\methodname combines professional interior assets and path-traced rendering across 12,081 scenes, 57.2\% of which belong to commercial or public-space semantic categories, with 2.16M panoramas carrying depth, normals, and exact poses along continuous trajectories (\cref{tab:dataset-comparison}).

%% file: sec/3_method_anpei.tex
\input{figures/pipeline}

\input{figures/lighting_repair}

\section{Building \methodname}
\label{sec:data}

We unify assets from 3ds Max, Unity, and other systems in Blender, assemble partial rooms, plan camera trajectories, refine lighting, and validate observations (\cref{fig:pipeline}). 
Reference-view ray probes assess viewpoint validity and geometric coverage, filtering out structurally incomplete sets. Complete scenes skip assembly.

\subsection{Multi-Room Scene Assembly}
\label{sec:houses}
\label{sec:preprocess}
\label{sec:complete}

\begingroup
\widowpenalty=10000
\looseness=-1
Many designer-authored assets depict a single room or a partial kitchen rather than a connected home.
We extract each room as a floor-aligned module with its candidate door openings and join modules by an in-plane rotation and translation that aligns opposing doors, so each room keeps its internal arrangement.\par
\endgroup

A constrained search avoids room overlaps and blocked doorways, favoring compact layouts connected through living and dining spaces.
We open designated wall segments, bridge floor gaps, and close unused boundaries, then verify traversability on the assembled geometry.

\subsection{Camera Trajectory Generation}
\label{sec:trajectory}
\label{sec:reach}

Authored cameras specify selected compositions rather than routes through an environment.
We build a navigation graph over the floor cells where a person-sized collision capsule fits and plan routes that balance spatial coverage and content visibility.
Coverage uses shortest walkable distance to the route, so nearby regions separated by walls do not count as visited; connected missed regions guide detours, including into small side rooms.
For content visibility, panoramas rendered at the authored cameras define the objects a walkthrough should see, and objects the route misses receive extra observation positions and short looks toward them.
We measure this visibility by casting rays from the moving perspective camera; the reference panoramas themselves do not count toward coverage.
The route becomes continuous motion with bounded turns and content-directed orientations, and views blocked by a wall or occluder are corrected by turning the camera without changing its path.
These orientations define default perspective views, while panoramas capture all directions at each sampled position.
For Sketchfab assets, a separate branch generates continuous 3D flight paths and evaluates surface coverage.

\subsection{Lighting Repair}
\label{sec:light}
\label{sec:lighting}

Lighting and material definitions differ across source systems, so conversion can change appearance and illumination in unintended ways.
After repairing incompatible materials and normalizing imported lights, we render panoramic previews at distributed trajectory positions.
We measure brightness per view and within six horizontal sectors of each panorama, which exposes dark corners that a scene-wide average hides, and use these measurements to calibrate exposure and place local fill lights.
An optional recovery branch rescales the energy of individual light groups within bounds, fitted jointly across views and regularized toward the authored lighting.
Each proposal is re-rendered and kept only if the photometric score improves and no view or sector becomes worse than before (\cref{fig:relight}).

\input{figures/modalities}
\input{figures/trajectory_stats}

\subsection{Scene Validation and Rendering}
\label{sec:audit}
\label{sec:agents}

Validation checks traversable connections, trajectory coverage and motion, and the usability of sampled renders across diverse viewpoints.
LLM-based agents diagnose failures from geometric reports and previews.
Policy agents revise shared processing rules across scenes, while per-scene agents propose structured local repair plans.
Deterministic executors apply these changes; a scene is accepted only after passing the checks that rejected it.
Acceptance criteria are provided in the supplementary material.

\label{sec:rendering}
We sample trajectories at 2\,Hz and path-trace $2048\times1024$ panoramas with aligned RGB, metric ray depth, world-space shading normals, and camera poses (\cref{fig:modalities}).
Scale follows recorded scene units or estimates from camera-height and architectural cues; the same conversion is applied to depth and translations.
Calibrated perspective views with different camera orientations and fields of view can be efficiently resampled without additional rendering.

Every rendered frame is audited for dark views and depth anomalies; diagnosed failures return to lighting or trajectory repair, with lighting rechecked after path changes.
Repairs are rerendered and re-audited under the same criteria, and persistent hard failures are excluded.

%% file: figures/pipeline.tex
\begin{figure*}[!t]
\centering
\resizebox{\textwidth}{!}{%
\begin{tikzpicture}[
  font=\sffamily,
  gatebox/.style={draw=memteal!70!black, fill=memteal!5, rounded corners=4pt, line width=0.7pt},
  mainbox/.style={draw=memteal!80!black, fill=memteal!10, rounded corners=4pt, line width=1.4pt},
  procbox/.style={draw=habiconv!80!black, fill=habiconv!6, rounded corners=4pt, line width=0.7pt},
  ttl/.style={font=\sffamily\bfseries\footnotesize, text=#1, anchor=base},
  arr/.style={-{Latex[length=2.4mm]}, line width=0.9pt, draw=mutedink},
  back/.style={-{Latex[length=2mm]}, line width=0.8pt, draw=red!65!black, dashed},
  backlbl/.style={font=\sffamily\small, text=red!65!black, fill=white, inner sep=1.5pt},
  qa/.style={-{Latex[length=2.2mm]}, line width=1.0pt, draw=viewblue!80!black, dashed},
  qalbl/.style={font=\sffamily\small, text=viewblue!80!black, fill=white, inner sep=1.5pt},
  wall/.style={line width=1.3pt, draw=black!70, line cap=round},
  furn/.style={fill=black!18, draw=none},
  ray/.style={line width=0.55pt, draw=viewblue!60},
  traj/.style={line width=1.55pt, draw=viewblue, -{Latex[length=1.9mm]}},
  look/.style={-{Latex[length=1.3mm]}, line width=0.65pt, draw=viewblue!65},
  zone/.style={draw=habinonres!85!black, dashed, line width=0.65pt},
  agent/.style={circle, fill=#1!85!black, text=white, font=\sffamily\bfseries\scriptsize, inner sep=0pt, minimum size=12pt},
  legendtext/.style={font=\sffamily\small, text=mutedink, anchor=west},
  modelimg/.style={inner sep=0pt},
  img/.style={inner sep=0pt, draw=white, line width=0.6pt},
  edge/.style={draw=mutedink!50, line width=0.35pt, line join=round},
]
\newcommand{\gbox}[4]{%
  \draw[#3] (#1-#2,-1.25) rectangle (#1+#2,1.35);
  \coordinate (#4w) at (#1-#2,0.05); \coordinate (#4e) at (#1+#2,0.05);}
\gbox{0}{1.35}{procbox}{src}
\node[modelimg] at (0,0.58) {\includegraphics[trim=133bp 119bp 188bp 85bp,clip,width=2.50cm,height=1.34cm,keepaspectratio]{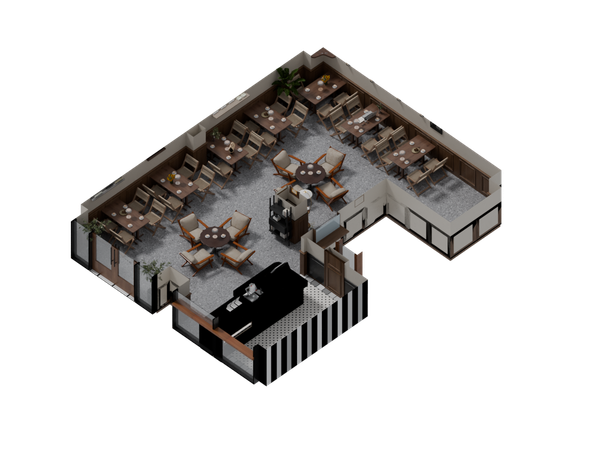}};
\node[modelimg] at (-0.65,-0.40) {\includegraphics[trim=76bp 74bp 128bp 39bp,clip,width=1.28cm,height=1.00cm,keepaspectratio]{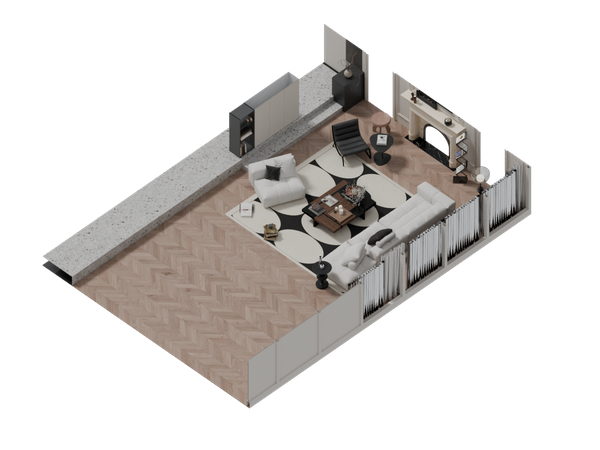}};
\node[modelimg] at (0.65,-0.40) {\includegraphics[trim=97bp 33bp 58bp 65bp,clip,width=1.28cm,height=1.00cm,keepaspectratio]{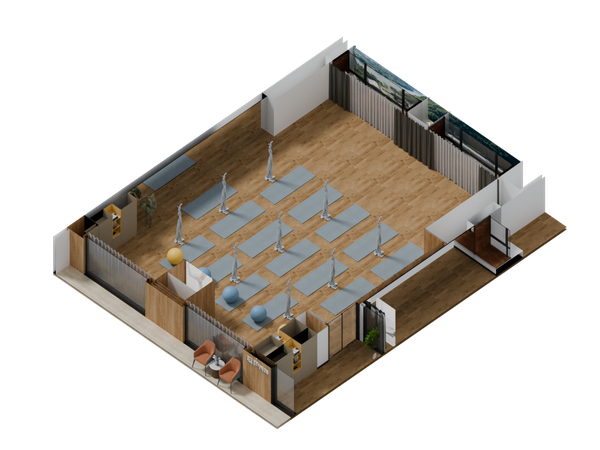}};
\node[ttl=mutedink] at (0,-1.02) {Sources};

\def\x{3.65}
\gbox{\x}{1.7}{gatebox}{pre}
\begin{scope}[shift={(\x,0.22)}]
  \draw[wall] (-1.05,-0.58) -- (-1.05,0.72) -- (1.05,0.72) -- (1.05,-0.58);
  \draw[red!65!black,dashed,line width=1.2pt] (-1.05,-0.58) -- (1.05,-0.58);
  \fill[furn] (0.48,0.35) rectangle (0.88,0.62);
  \foreach \a in {0,45,90,135,180} \draw[ray] (0,0) -- (\a:0.66);
  \foreach \a in {225,270,315} \draw[red!65!black,line width=0.65pt] (0,0) -- (\a:0.88);
  \fill[viewblue] (0,0) circle (2.2pt);
\end{scope}
\node[ttl=memteal!70!black] at (\x,-1.02) {Preprocessing};

\def\x{7.5}
\gbox{\x}{1.55}{gatebox}{hou}
\begin{scope}[shift={(\x,0.22)}]
  \fill[memteal!10] (0,-0.6) rectangle (1.08,0.72);
  \fill[furn] (-0.95,0.34) rectangle (-0.55,0.62);
  \fill[furn] (-0.48,-0.51) rectangle (-0.15,-0.28);
  \fill[furn] (0.45,0.29) rectangle (0.96,0.62);
  \fill[furn] (0.22,-0.51) rectangle (0.50,-0.32);
  \draw[wall] (-1.08,-0.6) rectangle (1.08,0.72);
  \draw[wall] (0,-0.6) -- (0,-0.09);
  \draw[wall] (0,0.30) -- (0,0.72);
  \fill[memteal!35] (-0.14,-0.09) rectangle (0.14,0.30);
  \draw[traj] plot[smooth,tension=0.65] coordinates
    {(-0.78,-0.34) (-0.73,0.06) (-0.32,0.24) (0,0.11) (0.35,-0.04) (0.78,-0.06) (0.81,-0.40)};
  \fill[habiartist] (-0.78,-0.34) circle (2.1pt);
\end{scope}
\node[ttl=memteal!70!black] at (\x,-1.02) {Multi-room assembly};

\def\x{11.85}
\gbox{\x}{2.2}{mainbox}{tra}
\begin{scope}[shift={(\x,0.22)}]
  \fill[memteal!8] (-0.6,-0.60) rectangle (0.6,0.72);
  \draw[wall] (-1.75,-0.60) rectangle (1.75,0.72);
  \foreach \xx in {-0.6,0.6} {
    \draw[wall] (\xx,-0.60) -- (\xx,-0.12);
    \draw[wall] (\xx,0.22) -- (\xx,0.72);
  }
  \fill[furn] (-1.55,0.39) rectangle (-0.90,0.61);
  \fill[furn] (-0.30,0.39) rectangle (0.24,0.61);
  \fill[furn] (0.92,0.35) rectangle (1.55,0.61);
  \draw[zone] (-0.43,0.25) rectangle (0.39,0.68);
  \draw[zone] (0.83,0.19) rectangle (1.65,0.68);
  \draw[traj] plot[smooth,tension=0.55] coordinates
    {(-1.48,-0.32) (-1.35,0.11) (-0.94,0.18) (-0.61,0.05) (-0.12,0.15) (0.30,-0.23) (0.61,0.04) (1.19,0.10) (1.49,-0.30)};
  \fill[habiartist] (-1.48,-0.32) circle (2.1pt);
  \draw[look] (-1.25,0.15) -- (-1.10,0.50);
  \draw[look] (-0.06,0.10) -- (0.02,0.46);
  \draw[look] (1.15,0.08) -- (1.32,0.47);
  \fill[black!80] (-1.50,0.09) -- (-1.66,-0.01) -- (-1.66,0.19) -- cycle;
  \fill[black!80] (1.51,0.05) -- (1.67,-0.05) -- (1.67,0.15) -- cycle;
\end{scope}
\node[ttl=memteal!80!black] at (\x,-1.02) {Trajectory generation};

\def\x{16.0}
\gbox{\x}{1.35}{procbox}{ren}
\node[img] at (\x+0.22,0.08) {\includegraphics[width=1.9cm,height=0.95cm]{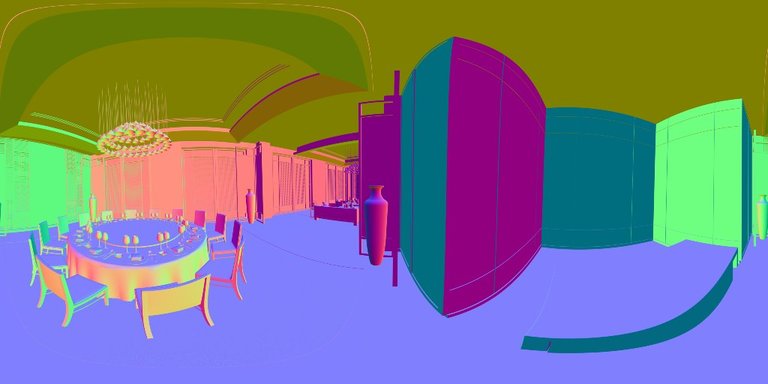}};
\node[img] at (\x,0.33) {\includegraphics[width=1.9cm,height=0.95cm]{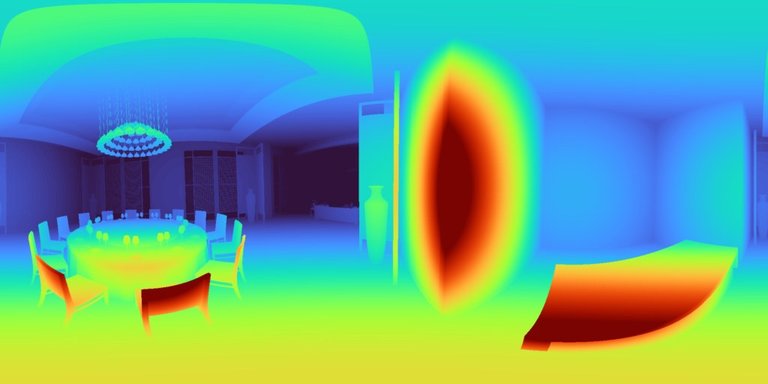}};
\node[img] at (\x-0.22,0.58) {\includegraphics[width=1.9cm,height=0.95cm]{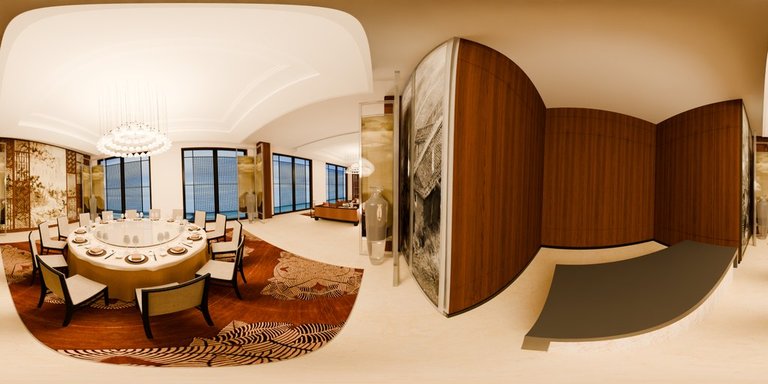}};
\node[ttl=mutedink] at (\x,-1.02) {Render};

\def\x{19.3}
\gbox{\x}{1.35}{gatebox}{aud}
\begin{scope}[shift={(\x,0.25)}]
  \foreach \r in {0,1} \foreach \c in {0,1,2} {
    \fill[viewblue!28] (-0.95+0.66*\c,0.08-0.55*\r) rectangle ++(0.56,0.42);
  }
  \fill[black!60] (-0.29,0.08) rectangle ++(0.56,0.42);
  \draw[red!65!black,line width=1pt] (-0.29,0.08) rectangle ++(0.56,0.42);
  \fill[white] (0.65,-0.25) circle (0.13);
  \draw[red!65!black,line width=1pt] (0.37,-0.47) rectangle ++(0.56,0.42);
\end{scope}
\node[ttl=memteal!70!black] at (\x,-1.02) {Frame audit};

\node[draw=memteal!70!black,fill=memteal!25,rounded corners=4pt,font=\sffamily\bfseries\small,
      minimum height=0.9cm,inner sep=5pt,anchor=west] (rel) at (21.25,0.05) {Release};
\draw[arr] (srce) -- (prew); \draw[arr] (pree) -- (houw); \draw[arr] (houe) -- (traw);
\draw[arr] (trae) -- (renw); \draw[arr] (rene) -- (audw); \draw[arr] (aude) -- (rel.west);
\draw[qa] (11.85,-1.25) -- (11.85,-1.65) -| (4.45,-1.25)
  node[qalbl,pos=0.25,below] {Lighting check on route panoramas};
\draw[back] (19.1,1.35) -- (19.1,1.8) -| (11.85,1.35)
  node[backlbl,pos=0.25,above] {depth hole: re-plan};
\draw[back] (19.7,1.35) -- (19.7,2.3) -| (4.45,1.35)
  node[backlbl,pos=0.25,above] {dark frame: relight};
\node[agent=habinonres] at (3.65+1.00,1.13) {P}; \node[agent=habihouse] at (3.65+1.48,1.13) {S};
\node[agent=habihouse] at (7.5+1.33,1.13) {S};
\node[agent=habinonres] at (11.85+1.98,1.13) {P};
\node[agent=habihouse] at (19.3+1.13,1.13) {S};

\node[agent=habinonres] at (0.50,-2.45) {P};
\node[legendtext] at (0.85,-2.45) {Policy agent};
\node[agent=habihouse] at (6.35,-2.45) {S};
\node[legendtext] at (6.70,-2.45) {Scene agent};
\draw[traj] (12.00,-2.45) -- (12.70,-2.45);
\node[legendtext] at (12.90,-2.45) {Camera route};
\draw[back] (18.20,-2.45) -- (18.90,-2.45);
\node[legendtext] at (19.10,-2.45) {Repair};
\end{tikzpicture}}
\vspace{-1.8em}
\caption{\textbf{\methodname is built around one continuous walk-through per scene.}
Light arrows, dashed boxes, and black triangles mark gaze directions, content zones, and camera anchors.
Lighting is checked on route panoramas; dark frames and depth holes trigger relighting and replanning, respectively.
Policy agents (P) update shared rules; scene agents (S) propose local repairs. All repairs are revalidated (\cref{sec:agents}).}
\label{fig:pipeline}
\end{figure*}

%% file: figures/lighting_repair.tex
\begin{figure}[t]
\centering
\includegraphics[width=\columnwidth]{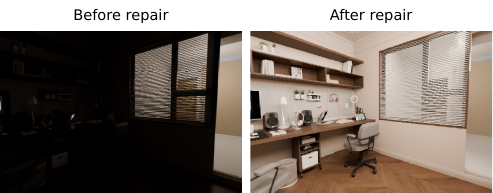}
\vspace{-0.6em}
\caption{\textbf{Lighting repair.} The same camera view before (left) and after (right) repair.}
\label{fig:relight}
\vspace{-0.6em}
\end{figure}

%% file: figures/modalities.tex
\begin{figure}[t]
\centering
\includegraphics[width=\columnwidth]{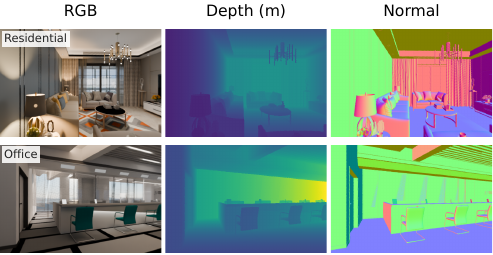}
\vspace{-0.6em}
\caption{\textbf{Aligned appearance and geometry.} RGB, metric camera-$z$ depth, and world-space shading normals at the same perspective viewpoint in a home (top) and an office (bottom). Both depth maps use the same 0--10\,m display range.}
\label{fig:modalities}
\vspace{-0.6em}
\end{figure}

%% file: figures/trajectory_stats.tex
\begin{figure}[t]
\centering
\includegraphics[width=\columnwidth]{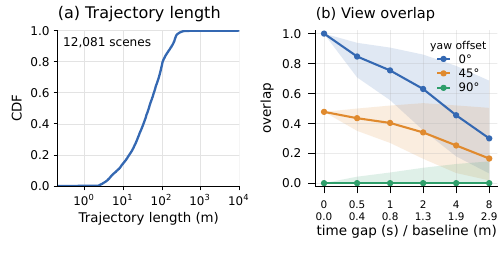}
\vspace{-1.8em}
\caption{\textbf{Trajectory statistics.}
(a) Cumulative distribution of per-scene camera path length across all \NumScenes scenes (logarithmic length axis).
(b) Overlap between two $90^\circ$ perspective views resampled from panoramas of the same trajectory, against the time gap (and median baseline) between the panoramas, for yaw offsets of $0^\circ$, $45^\circ$, and $90^\circ$ (median and interquartile range; computed over 1,000 independently sampled scenes).}
\label{fig:stats-main}
\vspace{-0.6em}
\end{figure}

%% file: sec/4_dataset_anpei.tex
\section{Dataset Analysis}
\label{sec:stats}

\subsection{Scene Composition}

\input{tables/category_counts}

\begin{figure}[t]
\centering
\includegraphics[width=\columnwidth]{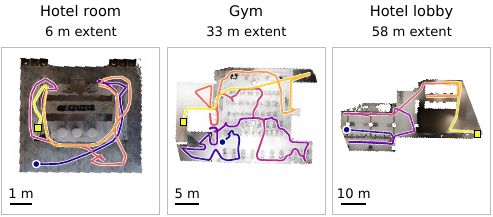}
\vspace{-0.6em}
\caption{\textbf{Walk-through trajectories} in three scenes of increasing extent, shown as top-down depth reconstructions with ceilings removed and dominant walls aligned. Path color progresses from purple to yellow over time (circle: start; square: end).}
\label{fig:trajectories}
\vspace{-0.6em}
\end{figure}

\input{tables/geometry_main}
\input{tables/nvs_main}

\methodname comprises \NumScenes scenes and \NumFramesExact panoramas, totaling \NumHours hours and \NumKm\,km of camera travel. It covers \NumMajor major and \NumMinor minor categories, spanning residential, commercial, and public spaces (\cref{tab:category-counts}). With approximately $26\times$ more scenes than Hypersim~\cite{data_hypersim}, it also provides diverse supervision and viewing configurations (\cref{tab:dataset-comparison}).

Residential spaces account for 38.5\% of scenes, while commercial and public spaces span 12 categories and comprise 6,907 scenes (57.2\%), highlighting broad coverage beyond residential environments.

Scene categories also vary in spatial extent and observation density. Transport trajectories span a median of 33.8\,m, compared with 7.4\,m for hospitality, indicating substantial differences in camera coverage. Transport also contributes 0.7\% of scenes but 1.3\% of panoramas, reflecting differences between scene-level and frame-level distributions.

\subsection{Trajectories and View Overlap}

A trajectory lasts a median of \MedDuration\,s and contains \MedFrames panoramas over \MedPath\,m of camera travel.
\cref{fig:stats-main}(a) shows the camera path-length distribution across all scenes, with equal weight per scene.
Assembled houses have a median path of 96\,m over a 29.6\,m extent.
We measure depth-verified overlap between $90^\circ$ perspective views resampled from the panoramas to quantify changes in observed content.

For views facing the same direction, median overlap falls from 85\% at a 0.5\,s gap to 63\% at 2\,s and 30\% at 8\,s, where the median baseline is 2.9\,m (\cref{fig:stats-main}(b)).
Wider temporal spacing reduces repeated observations but also reduces shared content for correspondence.
A $45^\circ$ yaw offset roughly halves overlap; at $90^\circ$, most pairs no longer overlap.
The yaw comparison uses the same camera positions, allowing orientation to change independently of translation baseline.
Multiple crops at one position add viewing directions, not translation baselines.
\cref{fig:trajectories} shows example trajectories across different extent of scenes.

\subsection{Geometric and Photometric Properties}

Depth reprojection gives a median relative error of \GeoMedRel over points passing the consistency test; the median panorama has 99.9\% valid depth coverage.
Visible surfaces are mostly nearby: \PDepthTwo of the sphere lies within 2\,m and \PDepthFive within 5\,m, even though trajectories can span much larger spaces.
After lighting repair, median measured luminance is \MedLum, with nearly black and clipped pixels each occupying less than 1\% of the median panorama.

%% file: tables/category_counts.tex
\begin{table}[t]
\centering
\captionsetup{skip=2pt}
\caption{\textbf{Scene and panorama counts by category.}}
\label{tab:category-counts}
\footnotesize
\setlength{\tabcolsep}{4pt}
\renewcommand{\arraystretch}{0.95}
\begin{tabular*}{\columnwidth}{@{\extracolsep{\fill}}lrr@{}}
\toprule
Category & \#Scenes & \#Panoramas (K) \\
\midrule
Residential & 4,657 & 901.4 \\
Hospitality & 3,691 & 544.3 \\
Leisure \& culture & 1,056 & 226.4 \\
Office & 601 & 127.2 \\
Food \& drink & 553 & 91.2 \\
Stylized / concept & 423 & 54.3 \\
Retail & 392 & 67.0 \\
Education & 233 & 42.2 \\
Sports \& fitness & 210 & 53.0 \\
Outdoor \& landscape & 94 & 12.2 \\
Transport & 88 & 27.9 \\
Public service & 44 & 5.0 \\
Personal service & 16 & 2.7 \\
Healthcare & 12 & 1.4 \\
Industrial & 11 & 2.5 \\
\midrule
\textbf{Total} & \textbf{\NumScenes} & \textbf{2,158.8} \\
\bottomrule
\end{tabular*}
\end{table}

%% file: tables/geometry_main.tex
\begin{table*}[!t]
\centering
\caption{\textbf{Quantitative results of multi-view geometry on \bench}.
Depth AbsRel and $\delta_1$ (\%) use per-scene median scaling; camera ATE (m) is the RMSE of camera centers after Sim(3) alignment; AUC is the pose AUC@30 (\%) over all view pairs.
Metrics are averaged over seeds within a scene and then equally over scenes; \textbf{bold} and \underline{underlined} denote the best and second best.}
\label{tab:bench-main}
\footnotesize
\setlength{\tabcolsep}{3.3pt}
\begin{tabular*}{\textwidth}{@{\extracolsep{\fill}}l*{12}{r}@{}}
\toprule
& \multicolumn{4}{c}{8 views} & \multicolumn{4}{c}{16 views} & \multicolumn{4}{c}{32 views} \\
\cmidrule(lr){2-5}\cmidrule(lr){6-9}\cmidrule(lr){10-13}
Method & AbsRel$\downarrow$ & $\delta_1\uparrow$ & ATE$\downarrow$ & AUC$\uparrow$ & AbsRel$\downarrow$ & $\delta_1\uparrow$ & ATE$\downarrow$ & AUC$\uparrow$ & AbsRel$\downarrow$ & $\delta_1\uparrow$ & ATE$\downarrow$ & AUC$\uparrow$ \\
\midrule
VGGT-$\Omega$~\cite{wang2026vggtomega} & \textbf{0.0438} & \textbf{97.3} & \textbf{0.133} & \textbf{93.8} & \textbf{0.0423} & \textbf{97.4} & \textbf{0.087} & \textbf{95.3} & \textbf{0.0431} & \textbf{97.4} & \textbf{0.149} & \textbf{95.4} \\
DA3-Giant 1.1~\cite{da3} & \underline{0.0474} & \underline{97.0} & 0.288 & 84.5 & \underline{0.0482} & \underline{97.1} & 0.250 & 86.5 & \underline{0.0535} & \underline{96.8} & \underline{0.206} & 87.3 \\
$\pi^3$~\cite{pi3} & 0.0649 & 96.8 & 0.338 & 86.4 & 0.0632 & 96.7 & 0.247 & 88.9 & 0.0687 & 96.4 & 0.283 & 89.2 \\
D\'ej\`a View~\cite{burzio2026dejaview} & 0.0662 & 95.6 & 0.370 & \underline{87.3} & 0.0681 & 95.8 & 0.280 & \underline{90.6} & 0.0739 & 95.6 & 0.263 & \underline{92.0} \\
WorldMirror 2.0~\cite{hyworld2026} & 0.0601 & 96.8 & 0.552 & 76.9 & 0.0648 & 96.8 & 0.387 & 82.4 & 0.0708 & 96.4 & 0.350 & 85.1 \\
MapAnything~\cite{keetha2026mapanything} & 0.1042 & 95.6 & \underline{0.284} & 83.8 & 0.1346 & 95.3 & \underline{0.226} & 85.2 & 0.1404 & 95.0 & 0.238 & 85.8 \\
R$^3$~\cite{xu2026r3} & 0.0674 & 95.0 & 0.881 & 63.6 & 0.0709 & 94.6 & 0.870 & 65.6 & 0.0727 & 94.6 & 0.674 & 70.2 \\
VGGT~\cite{vggt} & 0.0753 & 94.4 & 0.724 & 69.0 & 0.0840 & 94.1 & 0.576 & 72.9 & 0.0889 & 93.6 & 0.564 & 74.0 \\
OmniVGGT~\cite{peng2025omnivggt} & 0.0815 & 93.3 & 1.035 & 61.6 & 0.0895 & 93.1 & 1.021 & 63.2 & 0.1009 & 92.4 & 1.032 & 63.5 \\
\bottomrule
\end{tabular*}

\vspace{-0.4em}
\par\vspace{4pt}
\begingroup
\captionsetup{type=figure,position=bottom,skip=10pt}
\input{figures/geometry_qualitative_content}
\endgroup
\end{table*}

%% file: figures/geometry_qualitative_content.tex
\input{figures/geometry_local4_panels}
  \caption{\textbf{Qualitative results of multi-view geometry on \bench.}
Colored point clouds of the same desk region, reconstructed by each method from four RGB views and shown next to the ground truth (GT).
}
  \label{fig:geometry-qualitative}

%% file: figures/geometry_local4_panels.tex
\begingroup
\newcommand{\SceneTwentyPanel}[2]{%
  \begin{minipage}[t]{0.196\linewidth}
    \centering
    {\fontsize{8}{9}\selectfont #2\par}
    \vspace{1pt}
    #1
  \end{minipage}}
\noindent%
\par\vspace{1em}
\SceneTwentyPanel{\includegraphics[width=\linewidth]{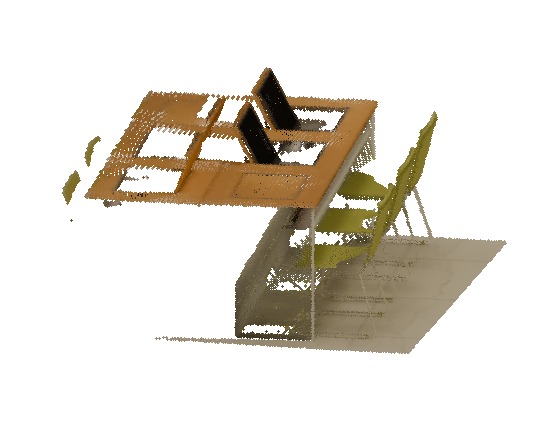}}{\textbf{GT}}\hfill%
\SceneTwentyPanel{\includegraphics[width=\linewidth]{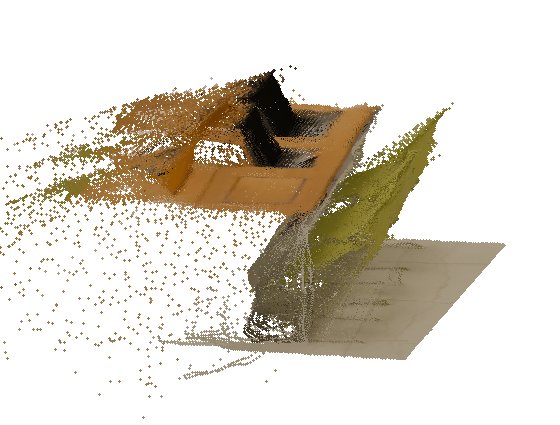}}{VGGT-$\Omega$}\hfill%
\SceneTwentyPanel{\includegraphics[width=\linewidth]{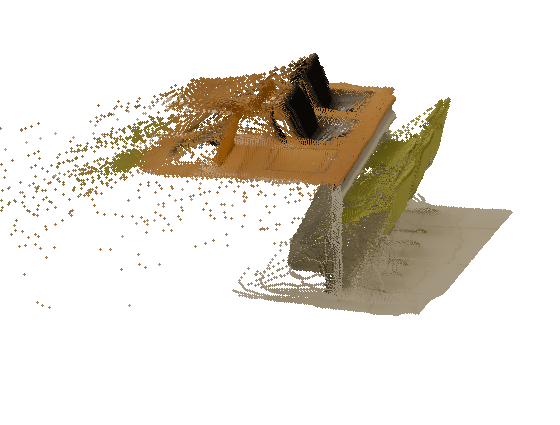}}{DA3-Giant}\hfill%
\SceneTwentyPanel{\includegraphics[width=\linewidth]{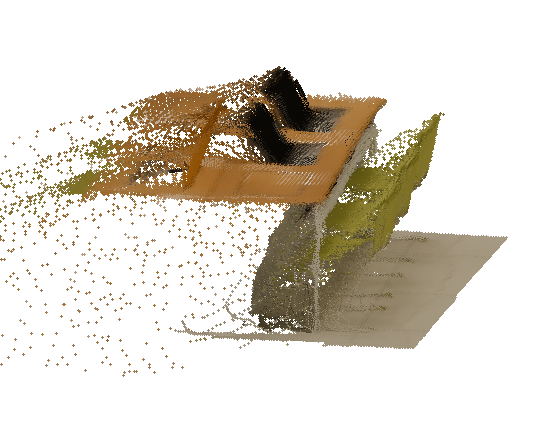}}{Pi3}\hfill%
\SceneTwentyPanel{\includegraphics[width=\linewidth]{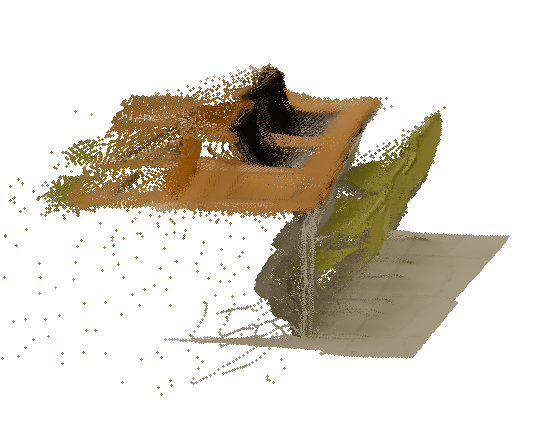}}{Deja-View}%
\par\vspace{3pt}
\noindent%
\SceneTwentyPanel{\includegraphics[width=\linewidth]{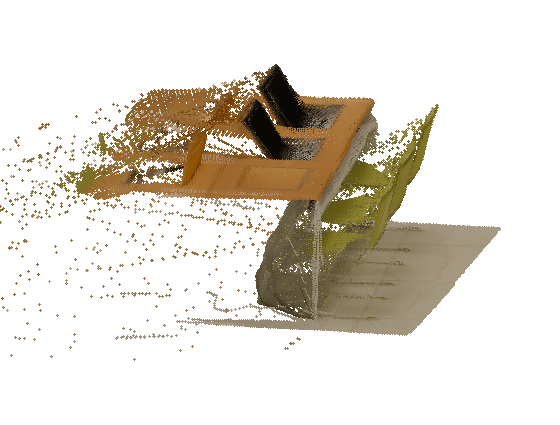}}{WorldMirror 2.0}\hfill%
\SceneTwentyPanel{\includegraphics[width=\linewidth]{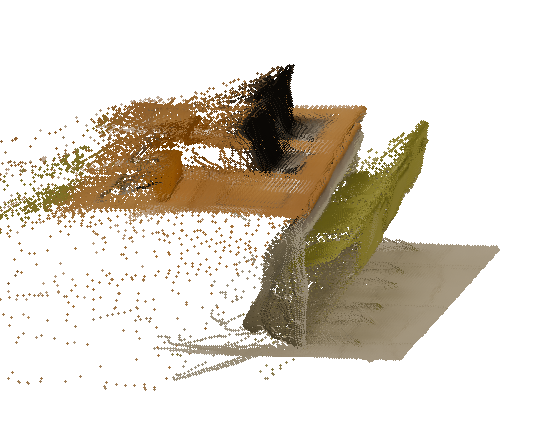}}{MapAnything}\hfill%
\SceneTwentyPanel{\includegraphics[width=\linewidth]{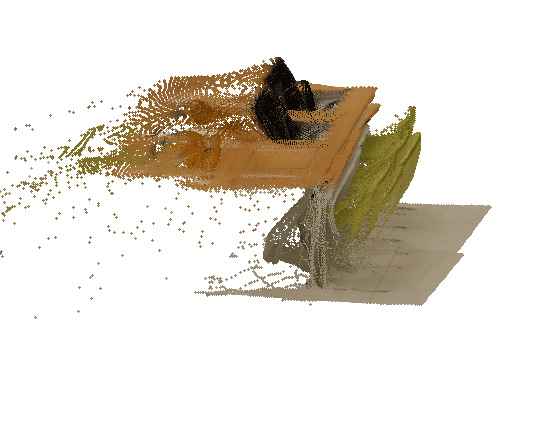}}{R3}\hfill%
\SceneTwentyPanel{\includegraphics[width=\linewidth]{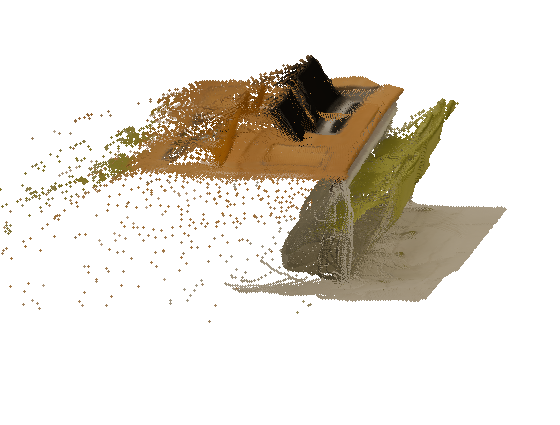}}{VGGT}\hfill%
\SceneTwentyPanel{\includegraphics[width=\linewidth]{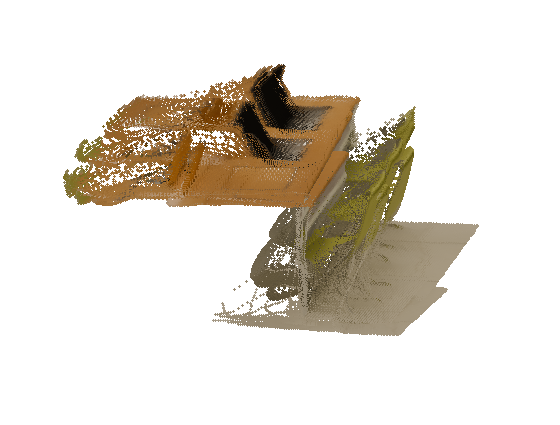}}{OmniVGGT}%
\par
\endgroup

%% file: tables/nvs_main.tex
\begin{table*}[!t]
\centering
\caption{\textbf{Quantitative results of feed-forward novel view synthesis on \bench.} 
``Camera'' lists the ground-truth camera information given to reconstruction (K: intrinsics, P: poses); repeated methods use different camera-input configurations. Gaussian frames are registered using context cameras only, and all targets are rendered with GT cameras. Groups separate configurations with and without GT context poses; rows within each group are sorted by PSNR at $N=16$.
\textbf{Bold} and \underline{underlined} mark the best and second best within each pose-input group. LPIPS uses the VGG network for all evaluated methods.}
\label{tab:nvs}
\footnotesize
\setlength{\tabcolsep}{3.2pt}
\resizebox{\textwidth}{!}{%
\begin{tabular}{@{}l c ccc ccc ccc ccc@{}}
\toprule
& & \multicolumn{3}{c}{$N=2$} & \multicolumn{3}{c}{$N=4$} & \multicolumn{3}{c}{$N=8$} & \multicolumn{3}{c}{$N=16$} \\
\cmidrule(lr){3-5}\cmidrule(lr){6-8}\cmidrule(lr){9-11}\cmidrule(lr){12-14}
Method & Camera & PSNR$\uparrow$ & SSIM$\uparrow$ & LPIPS$\downarrow$ & PSNR$\uparrow$ & SSIM$\uparrow$ & LPIPS$\downarrow$ & PSNR$\uparrow$ & SSIM$\uparrow$ & LPIPS$\downarrow$ & PSNR$\uparrow$ & SSIM$\uparrow$ & LPIPS$\downarrow$ \\
\midrule
ReSplat~\cite{xu2026resplat} & K, P & \textbf{21.73} & \textbf{0.762} & 0.373 & \textbf{24.27} & \textbf{0.809} & 0.308 & \textbf{25.97} & \textbf{0.842} & 0.263 & \textbf{27.40} & \textbf{0.865} & \underline{0.232} \\
LVSM~\cite{jin2025lvsm} & K, P & 20.46 & 0.732 & \underline{0.318} & 22.37 & 0.772 & \textbf{0.269} & \underline{23.88} & 0.800 & \textbf{0.232} & \underline{24.42} & 0.808 & \textbf{0.221} \\
SplatAnything~\cite{yu2026splatanything} & K, P & 20.26 & \underline{0.740} & \textbf{0.314} & 22.40 & \underline{0.791} & \textbf{0.269} & 23.50 & \underline{0.812} & \underline{0.240} & 24.30 & \underline{0.826} & \textbf{0.221} \\
DA3-Giant 1.1 (GS)~\cite{da3} & K, P & 15.76 & 0.599 & 0.500 & 20.07 & 0.728 & 0.359 & 21.87 & 0.760 & 0.299 & 23.00 & 0.772 & 0.267 \\
WorldMirror 2.0 (GS)~\cite{hyworld2026} & K, P & 18.40 & 0.665 & 0.367 & 20.86 & 0.739 & 0.313 & 22.03 & 0.768 & 0.276 & 22.84 & 0.788 & 0.255 \\
DepthSplat~\cite{xu2025depthsplat} & K, P & \underline{21.18} & 0.737 & 0.333 & \underline{22.80} & \underline{0.791} & \underline{0.278} & 22.57 & 0.802 & 0.265 & 21.45 & 0.787 & 0.283 \\
YoNoSplat~\cite{ye2026yonosplat} & K, P & 17.80 & 0.649 & 0.462 & 19.51 & 0.685 & 0.419 & 19.97 & 0.693 & 0.400 & 20.17 & 0.698 & 0.396 \\
YoNoSplat~\cite{ye2026yonosplat} & P & 15.71 & 0.600 & 0.512 & 17.21 & 0.634 & 0.482 & 17.60 & 0.641 & 0.471 & 17.83 & 0.646 & 0.470 \\
\midrule
SplatAnything~\cite{yu2026splatanything} & -- & \textbf{18.00} & \textbf{0.648} & \textbf{0.371} & \textbf{19.55} & \underline{0.680} & \textbf{0.332} & \textbf{20.67} & \underline{0.709} & \textbf{0.298} & \textbf{21.75} & \textbf{0.736} & \textbf{0.267} \\
DA3-Giant 1.1 (GS)~\cite{da3} & -- & 16.13 & \underline{0.613} & 0.504 & \underline{19.37} & \textbf{0.704} & 0.400 & \underline{20.42} & \textbf{0.721} & 0.357 & \underline{21.27} & \underline{0.730} & 0.329 \\
WorldMirror 2.0 (GS)~\cite{hyworld2026} & -- & \underline{16.60} & 0.580 & \underline{0.424} & 18.89 & 0.653 & \underline{0.370} & 20.08 & 0.686 & \underline{0.328} & 21.01 & 0.711 & \underline{0.301} \\
AnySplat~\cite{jiang2025anysplat} & -- & 13.65 & 0.471 & 0.531 & 15.92 & 0.554 & 0.464 & 16.97 & 0.593 & 0.424 & 17.71 & 0.614 & 0.397 \\
YoNoSplat~\cite{ye2026yonosplat} & -- & 15.61 & 0.600 & 0.510 & 16.94 & 0.627 & 0.477 & 17.37 & 0.638 & 0.458 & 17.58 & 0.643 & 0.453 \\
E-RayZer~\cite{zhao2026erayzer} & -- & 12.02 & 0.527 & 0.571 & 14.61 & 0.587 & 0.545 & 15.22 & 0.602 & 0.525 & 15.11 & 0.607 & 0.531 \\
\bottomrule
\end{tabular}}

\par\vspace{4pt}
\begingroup
\captionsetup{type=figure,position=bottom,skip=10pt}
\input{figures/nvs_qualitative_content}
\endgroup
\end{table*}

%% file: figures/nvs_qualitative_content.tex
\includegraphics[width=\textwidth]{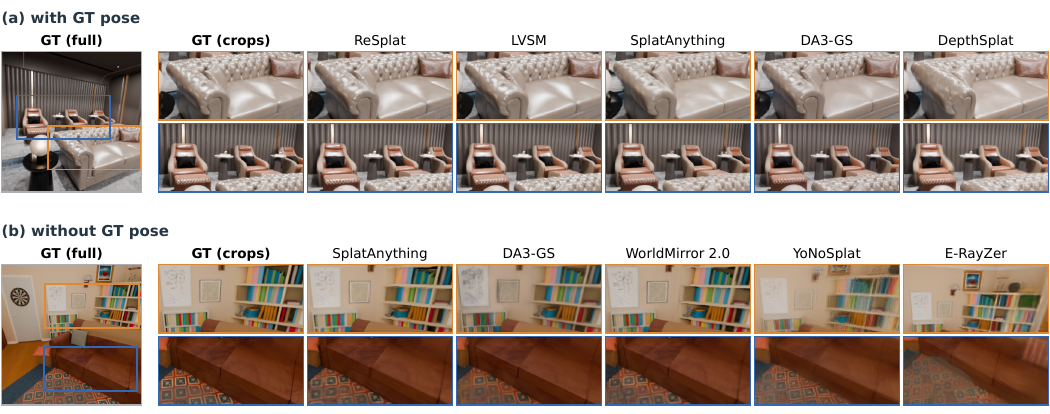}
\vspace{-1.6em}
\caption{\textbf{Qualitative results of feed-forward novel view synthesis on \bench.}
Each setting uses eight context images and shows the full GT target on the left, followed by two matched local crops for GT and each method.
Orange and blue boxes on the full GT image mark the regions shown in the corresponding crop rows.
(a) With GT pose uses the K, P configurations in \cref{tab:nvs}; (b) without GT pose uses RGB only.
GT target cameras are used for rendering in both settings.}
\label{fig:nvs-qualitative}
\vspace{-0.6em}

%% file: sec/5_experiments_anpei.tex
\section{\bench}

\label{sec:experiments}

The realistic scenes and accurate annotations in \methodname allow us to benchmark how well current 3D models reconstruct indoor environments.
We evaluate feed-forward multi-view geometry and novel view synthesis (NVS) zero-shot on held-out scenes of \methodname.

\subsection{Setup}
\label{sec:bench-setup}

\paragraph{Scenes and views.}
\bench contains 117 evaluation scenes spanning eight semantic categories: its production groups comprise 41 residential rooms, 69 non-residential venues, six Sketchfab models, and one assembled house.
For each trajectory, we resample the panoramas into $512\times512$ perspective views oriented along the camera's forward direction, with RGB, camera-$z$ depth, normals, and exact camera parameters.
View selection uses ground-truth covisibility: geometry uses connected, nested sets of 8, 16, and 32 views for three seeds; NVS uses nested context sets with $N\in\{2,4,8,16\}$ and four disjoint target views per scene.
Each NVS target is at least 20\% visible from each context view and 80\% covered by their union.
See supplementary material for selection thresholds and model settings.

\paragraph{Metrics and baselines.}
For geometry, we report depth AbsRel and $\delta_1$ after per-scene median scaling, the camera absolute trajectory error (ATE) after Sim(3) alignment~\cite{umeyama1991least}, and the pose AUC@30 over all view pairs.
For NVS, we report PSNR, SSIM~\cite{wang2004ssim}, and LPIPS~\cite{zhang2018lpips}.
We evaluate 9 geometry models from RGB input alone and 9 NVS models, additionally testing 4 NVS models with their native GT-context-camera interfaces. The context-camera information is listed for each configuration; predicted Gaussian frames are aligned using only context cameras before target rendering.
Metrics are averaged within scenes over seeds or targets, then equally over scenes.

\subsection{Multi-View Geometry}
\label{sec:bench-results}

\paragraph{Depth Accuracy.}
VGGT-$\Omega$ achieves the lowest depth AbsRel and highest $\delta_1$ at every view budget, with an AbsRel of 0.042 at 16 views (\cref{tab:bench-main}).
All methods exceed 92\% in $\delta_1$ after median scaling.
Adding views raises the depth error averaged over all input views for most methods.
On the same eight anchor views, however, depth error decreases for all nine methods as context grows, with depth scaling fitted only on those anchors.
For example, DA3's anchor AbsRel falls from 0.0474 to 0.0440 between 8 and 32 views, while its full-set AbsRel rises to 0.0535.
Additional context improves predictions on the original views even as newly scored views raise the full-set error.
The two scores distinguish the benefit of more context from changes in which views are evaluated.

\paragraph{Camera Pose Accuracy.}
\looseness=-1
VGGT-$\Omega$ achieves the lowest ATE and highest pose AUC at every view budget (\cref{tab:bench-main}).
At 16 views, ATE ranges from 0.087 to 1.021\,m, with MapAnything ranking second at 0.226\,m.
Increasing the input budget from 8 to 32 views improves pose AUC for every method, although ATE does not decrease monotonically for all methods.
From the most compact to the most extended fifth of the view sets, median extent increases from 3.1 to 13.7\,m (about $4.4\times$).
Across these groups, median ATE grows by $4.2\times$ for VGGT-$\Omega$ and up to $17\times$ for OmniVGGT, while the pose AUC of the four strongest methods changes by at most 5.2 points.
Non-residential venues also have higher ATE than residential rooms, by $2.7\times$ for VGGT-$\Omega$ and $3.7\times$ for DA3.
These results motivate reporting extent alongside scene function.

\cref{fig:geometry-qualitative} compares local colored point clouds reconstructed from four RGB views of the same desk region.
A shared side-view direction, distinct from all input views, exposes misaligned surfaces and duplicated edges around the desk, screens, and chairs.
All methods share the same viewing direction, scale, and crop.

\subsection{Novel View Synthesis}
\label{sec:nvs}

\paragraph{Posed NVS.}
With GT context poses, ReSplat achieves the highest PSNR and SSIM at every $N$, with PSNR improving from 21.73 to 27.40\,dB between two and 16 views (\cref{tab:nvs}).
LVSM and camera-conditioned SplatAnything attain the lowest LPIPS across the budgets, so the best pixel fidelity and perceptual similarity occur in different configurations.
DepthSplat's PSNR peaks at four views, while its SSIM and LPIPS are best at eight.
These configurations also receive GT intrinsics, except for the poses-only YoNoSplat variant.
At $N=4$, supplying intrinsics in addition to poses raises YoNoSplat's PSNR from 17.21 to 19.51\,dB, further demonstrating that known camera poses alone do not fully remove the effect of intrinsic calibration.

\paragraph{Unposed NVS.}
Without GT context poses, SplatAnything has the highest reported PSNR and lowest LPIPS at each budget, reaching 21.75\,dB and 0.267 at 16 views; E-RayZer's PSNR remains near 15\,dB at eight and 16 views.
To assess the effect of camera inputs within these models, we compare their posed and unposed configurations with the checkpoint and input resolution held fixed.
Supplying GT intrinsics and poses raises SplatAnything's PSNR from 21.75 to 24.30\,dB at $N=16$.
For YoNoSplat at $N=4$, GT poses alone add only 0.27\,dB over the unposed result, whereas poses and intrinsics together add 2.57\,dB.
DA3 also benefits at 16 views, improving from 21.27 to 23.00\,dB, but its PSNR drops from 16.13 to 15.76\,dB at two views.
The effect of camera conditioning varies across models and view budgets.

\cref{fig:nvs-qualitative} gives two examples at eight context views, clearly separating GT context cameras from RGB-only reconstruction results. Matched local crops further show subtle differences in sofa upholstery and bookshelf edges.

%% file: sec/6_conclusion.tex
\section{Conclusion}
\label{sec:conclusion}

We introduced \methodname, a dataset of \NumScenes professionally authored indoor scenes and \NumFrames path-traced panoramas with aligned RGB, metric ray depth, shading normals, and camera poses.
Our agent-in-the-loop Blender pipeline combines room assembly, trajectory planning, lighting repair, and deterministic validation.
\bench evaluates camera estimation, depth prediction, and novel view synthesis across diverse interiors.
Challenges remain in camera estimation over extended spaces and novel view synthesis without known input poses.
We will release the renderings, per-scene metadata, and benchmark.

%% file: sec/7_ack.tex
\section*{Acknowledgments}
This work was supported by the National Natural Science Foundation of China (Grant No.~62606405). We thank AlilangArt for generously sharing 1,500 scenes with us.
We also acknowledge the use of Codex and Claude Code to assist with dataset construction and improve the language, grammar, and readability of the manuscript.

%% file: habi.bib
@inproceedings{vggt,
  title     = {{VGGT}: Visual Geometry Grounded Transformer},
  author    = {Wang, Jianyuan and Chen, Minghao and Karaev, Nikita and Vedaldi, Andrea and Rupprecht, Christian and Novotny, David},
  booktitle = {CVPR},
  year      = {2025}
}

@inproceedings{keetha2026mapanything,
  title     = {{MapAnything}: Universal Feed-Forward Metric {3D} Reconstruction},
  author    = {Nikhil Keetha and Norman M\"{u}ller and Johannes Sch\"{o}nberger and Lorenzo Porzi and Yuchen Zhang and Tobias Fischer and Arno Knapitsch and Duncan Zauss and Ethan Weber and Nelson Antunes and Jonathon Luiten and Manuel Lopez-Antequera and Samuel Rota Bul\`{o} and Christian Richardt and Deva Ramanan and Sebastian Scherer and Peter Kontschieder},
  booktitle = {3DV},
  year      = {2026}
}

@inproceedings{yeshwanth2023scannetpp,
  title     = {{ScanNet++}: A High-Fidelity Dataset of {3D} Indoor Scenes},
  author    = {Yeshwanth, Chandan and Liu, Yueh-Cheng and Nie{\ss}ner, Matthias and Dai, Angela},
  booktitle = {ICCV},
  year      = {2023}
}

@inproceedings{pi3,
  title     = {{$\pi^3$: Permutation-Equivariant Visual Geometry Learning}},
  author    = {Wang, Yifan and Zhou, Jianjun and Zhu, Haoyi and Chang, Wenzheng and Zhou, Yang and Li, Zizun and Chen, Junyi and Pang, Jiangmiao and Shen, Chunhua and He, Tong},
  booktitle = {ICLR},
  year      = {2026}
}

@inproceedings{da3,
  title     = {{Depth Anything 3: Recovering the Visual Space from Any Views}},
  author    = {Lin, Haotong and Chen, Sili and Liew, Jun Hao and Chen, Donny Y. and Li, Zhenyu and Zhao, Yang and Peng, Sida and Guo, Hengkai and Zhou, Xiaowei and Shi, Guang and Feng, Jiashi and Kang, Bingyi},
  booktitle = {ICLR},
  year      = {2026}
}

@inproceedings{data_dl3dv,
  title     = {{DL3DV-10K}: A Large-Scale Scene Dataset for Deep Learning-based {3D} Vision},
  author    = {Ling, Lu and Sheng, Yichen and Tu, Zhi and Zhao, Wentian and Xin, Cheng and Wan, Kun and Yu, Lantao and Guo, Qianyu and Yu, Zixun and Lu, Yawen and Li, Xuanmao and Sun, Xingpeng and Ashok, Rohan and Mukherjee, Aniruddha and Kang, Hao and Kong, Xiangrui and Hua, Gang and Zhang, Tianyi and Benes, Bedrich and Bera, Aniket},
  booktitle = {CVPR},
  year      = {2024}
}

@inproceedings{data_hypersim,
  title     = {{{Hypersim}: {A} Photorealistic Synthetic Dataset for Holistic Indoor Scene Understanding}},
  author    = {Mike Roberts AND Jason Ramapuram AND Anurag Ranjan AND Atulit Kumar AND
                 Miguel Angel Bautista AND Nathan Paczan AND Russ Webb AND Joshua M. Susskind},
  booktitle = {ICCV},
  year      = {2021}
}

@inproceedings{data_scannet,
  title     = {{ScanNet: Richly-annotated 3D Reconstructions of Indoor Scenes}},
  author    = {Dai, Angela and Chang, Angel X. and Savva, Manolis and Halber, Maciej and Funkhouser, Thomas and Nie{\ss}ner, Matthias},
  booktitle = {CVPR},
  year      = {2017}
}

@inproceedings{data_arkitscenes,
  title     = {{ARKitScenes}: A Diverse Real-World Dataset for {3D} Indoor Scene Understanding Using Mobile {RGB-D} Data},
  author    = {Gilad Baruch and Zhuoyuan Chen and Afshin Dehghan and Tal Dimry and Yuri Feigin and Peter Fu and Thomas Gebauer and Brandon Joffe and Daniel Kurz and Arik Schwartz and Elad Shulman},
  booktitle = {NeurIPS},
  year      = {2021}
}

@inproceedings{data_ase,
  title     = {{SceneScript}: Reconstructing Scenes with an Autoregressive Structured Language Model},
  author    = {Avetisyan, Armen and Xie, Christopher and Howard-Jenkins, Henry and Yang, Tsun-Yi and Aroudj, Samir and Patra, Suvam and Zhang, Fuyang and Holland, Luke and Frost, Duncan and Orme, Campbell and Engel, Jakob and Miller, Edward and Newcombe, Richard and Balntas, Vasileios},
  booktitle = {ECCV},
  year      = {2024}
}

@article{armeni2017joint,
  title     = {Joint {2D-3D}-Semantic Data for Indoor Scene Understanding},
  author    = {Armeni, Iro and Sax, Sasha and Zamir, Amir R. and Savarese, Silvio},
  journal   = {arXiv},
  year      = {2017}
}

@inproceedings{chang2017matterport3d,
  title     = {{Matterport3D}: Learning from {RGB-D} Data in Indoor Environments},
  author    = {Chang, Angel and Dai, Angela and Funkhouser, Thomas and Halber, Maciej and Nie{\ss}ner, Matthias and Savva, Manolis and Song, Shuran and Zeng, Andy and Zhang, Yinda},
  booktitle = {3DV},
  year      = {2017}
}

@article{straub2019replica,
  title     = {The {Replica} Dataset: A Digital Replica of Indoor Spaces},
  author    = {Straub, Julian and Whelan, Thomas and Ma, Lingni and Chen, Yufan and Wijmans, Erik and Green, Simon and Engel, Jakob J. and Mur-Artal, Raul and Ren, Carl and Verma, Shobhit and Clarkson, Anton and Yan, Mingfei and Budge, Brian and Yan, Yajie and Pan, Xiaqing and Yon, June and Zou, Yuyang and Leon, Kimberly and Carter, Nigel and Briales, Jesus and Gillingham, Tyler and Mueggler, Elias and Pesqueira, Luis and Savva, Manolis and Batra, Dhruv and Strasdat, Hauke M. and De Nardi, Renzo and Goesele, Michael and Lovegrove, Steven and Newcombe, Richard},
  journal   = {arXiv},
  year      = {2019}
}

@inproceedings{zheng2020structured3d,
  title     = {{Structured3D}: A Large Photo-realistic Dataset for Structured {3D} Modeling},
  author    = {Zheng, Jia and Zhang, Junfei and Li, Jing and Tang, Rui and Gao, Shenghua and Zhou, Zihan},
  booktitle = {ECCV},
  year      = {2020}
}

@inproceedings{fu20213dfront,
  title     = {{3D-FRONT}: {3D} Furnished Rooms with Layouts and Semantics},
  author    = {Fu, Huan and Cai, Bowen and Gao, Lin and Zhang, Ling-Xiao and Wang, Jiaming and Li, Cao and Zeng, Qixun and Sun, Chengyue and Jia, Rongfei and Zhao, Binqiang and Zhang, Hao},
  booktitle = {ICCV},
  year      = {2021}
}

@inproceedings{li2018interiornet,
  title     = {{InteriorNet}: Mega-scale Multi-sensor Photo-realistic Indoor Scenes Dataset},
  author    = {Li, Wenbin and Saeedi, Sajad and McCormac, John and Clark, Ronald and Tzoumanikas, Dimos and Ye, Qing and Huang, Yuzhong and Tang, Rui and Leutenegger, Stefan},
  booktitle = {BMVC},
  year      = {2018}
}

@inproceedings{deitke2022procthor,
  title     = {{ProcTHOR}: Large-Scale Embodied {AI} Using Procedural Generation},
  author    = {Deitke, Matt and VanderBilt, Eli and Herrasti, Alvaro and Weihs, Luca and Salvador, Jordi and Ehsani, Kiana and Han, Winson and Kolve, Eric and Farhadi, Ali and Kembhavi, Aniruddha and Mottaghi, Roozbeh},
  booktitle = {NeurIPS},
  year      = {2022}
}

@inproceedings{raistrick2024infinigenindoors,
  title     = {{Infinigen Indoors}: Photorealistic Indoor Scenes using Procedural Generation},
  author    = {Raistrick, Alexander and Mei, Lingjie and Kayan, Karhan and Yan, David and Zuo, Yiming and Han, Beining and Wen, Hongyu and Parakh, Meenal and Alexandropoulos, Stamatis and Lipson, Lahav and Ma, Zeyu and Deng, Jia},
  booktitle = {CVPR},
  year      = {2024}
}

@inproceedings{xia2018gibson,
  title     = {{Gibson Env}: Real-World Perception for Embodied Agents},
  author    = {Xia, Fei and Zamir, Amir R. and He, Zhiyang and Sax, Alexander and Malik, Jitendra and Savarese, Silvio},
  booktitle = {CVPR},
  year      = {2018}
}

@inproceedings{ramakrishnan2021hm3d,
  title     = {{Habitat-Matterport} {3D} Dataset ({HM3D}): 1000 Large-scale {3D} Environments for Embodied {AI}},
  author    = {Ramakrishnan, Santhosh K. and Gokaslan, Aaron and Wijmans, Erik and Maksymets, Oleksandr and Clegg, Alex and Turner, John and Undersander, Eric and Galuba, Wojciech and Westbury, Andrew and Chang, Angel X. and Savva, Manolis and Zhao, Yili and Batra, Dhruv},
  booktitle = {NeurIPS},
  year      = {2021}
}

@article{umeyama1991least,
  title     = {Least-Squares Estimation of Transformation Parameters Between Two Point Patterns},
  author    = {Umeyama, Shinji},
  journal   = {IEEE TPAMI},
  year      = {1991}
}

@inproceedings{xu2025depthsplat,
  title     = {{DepthSplat}: Connecting {G}aussian Splatting and Depth},
  author    = {Xu, Haofei and Peng, Songyou and Wang, Fangjinhua and Blum, Hermann and Barath, Daniel and Geiger, Andreas and Pollefeys, Marc},
  booktitle = {CVPR},
  year      = {2025}
}

@inproceedings{xu2026resplat,
  title     = {{ReSplat}: Learning Recurrent {G}aussian Splatting},
  author    = {Xu, Haofei and Barath, Daniel and Geiger, Andreas and Pollefeys, Marc},
  booktitle = {ECCV},
  year      = {2026}
}

@inproceedings{ye2026yonosplat,
  title     = {{YoNoSplat}: You Only Need One Model for Feedforward {3D} {G}aussian Splatting},
  author    = {Ye, Botao and Chen, Boqi and Xu, Haofei and Barath, Daniel and Pollefeys, Marc},
  booktitle = {ICLR},
  year      = {2026}
}

@article{wang2004ssim,
  title     = {Image Quality Assessment: From Error Visibility to Structural Similarity},
  author    = {Wang, Zhou and Bovik, Alan C. and Sheikh, Hamid R. and Simoncelli, Eero P.},
  journal   = {IEEE TIP},
  year      = {2004}
}

@inproceedings{zhang2018lpips,
  title     = {The Unreasonable Effectiveness of Deep Features as a Perceptual Metric},
  author    = {Zhang, Richard and Isola, Phillip and Efros, Alexei A. and Shechtman, Eli and Wang, Oliver},
  booktitle = {CVPR},
  year      = {2018}
}

@article{jiang2025anysplat,
  title     = {{AnySplat}: Feed-forward {3D} {G}aussian Splatting from Unconstrained Views},
  author    = {Jiang, Lihan and Mao, Yucheng and Xu, Linning and Lu, Tao and Ren, Kerui and Jin, Yichen and Xu, Xudong and Yu, Mulin and Pang, Jiangmiao and Zhao, Feng and Lin, Dahua and Dai, Bo},
  journal   = {ACM TOG},
  year      = {2025}
}

@article{hyworld2026,
  title     = {{HY-World 2.0}: A Multi-Modal World Model for Reconstructing, Generating, and Simulating {3D} Worlds},
  author    = {{Team HY-World} and Cao, Chenjie and Zuo, Xuhui and Wang, Zhenwei and Zhang, Yisu and Wu, Junta and Liu, Zhenyang and Gong, Yuning and Liu, Yang and Yuan, Bo and Zhang, Chao and Li, Coopers and Guo, Dongyuan and Yang, Fan and Zhang, Haiyu and Cao, Hang and Zhu, Jianchen and Lin, Jiaxin and Xiao, Jie and Zhang, Jihong and Yu, Junlin and Wang, Lei and Wang, Lifu and Wang, Lilin and {Linus} and Chen, Minghui and He, Peng and Zhao, Penghao and Chen, Qi and Chen, Rui and Shao, Rui and Liu, Sicong and Qin, Wangchen and Niu, Xiaochuan and Yuan, Xiang and Sun, Yi and Tang, Yifei and Sun, Yifu and Lian, Yihang and Tan, Yonghao and Liu, Yuhong and Yin, Yuyang and Min, Zhiyuan and Wang, Tengfei and Guo, Chunchao},
  journal   = {arXiv},
  year      = {2026}
}

@inproceedings{wang2026vggtomega,
  title     = {{VGGT}-{$\Omega$}},
  author    = {Wang, Jianyuan and Chen, Minghao and Zhang, Shangzhan and Karaev, Nikita and Sch{\"o}nberger, Johannes and Labatut, Patrick and Bojanowski, Piotr and Novotny, David and Vedaldi, Andrea and Rupprecht, Christian},
  booktitle = {CVPR},
  year      = {2026}
}

@inproceedings{burzio2026dejaview,
  title     = {{D\'ej\`a View}: Looping Transformers for Multi-View {3D} Reconstruction},
  author    = {Burzio, Alessandro and Fischer, Tobias and Elflein, Sven and Zhou, Qunjie and de Lutio, Riccardo and Ren, Jiawei and Huang, Jiahui and Huang, Shengyu and Pollefeys, Marc and Leal-Taix{\'e}, Laura and Gojcic, Zan and Turki, Haithem},
  booktitle = {NeurIPS},
  year      = {2026}
}

@inproceedings{xu2026r3,
  title     = {{$R^3$}: {3D} Reconstruction via Relative Regression},
  author    = {Xu, Congrong and Gao, Huachen and Chen, Xingyu and Xiu, Yuliang and Gao, Jun and Chen, Anpei},
  booktitle = {NeurIPS},
  year      = {2026}
}

@inproceedings{peng2025omnivggt,
  title     = {{OmniVGGT}: Omni-Modality Driven Visual Geometry Grounded Transformer},
  author    = {Peng, Haosong and Li, Hao and Dai, Yalun and Lan, Yushi and Luo, Yihang and Qi, Tianyu and Zhang, Zhengshen and Zhan, Yufeng and Zhang, Junfei and Xu, Wenchao and Liu, Ziwei},
  booktitle = {CVPR},
  year      = {2026}
}

@inproceedings{mccormac2017scenenet,
  title     = {{SceneNet RGB-D}: Can {5M} Synthetic Images Beat Generic {ImageNet} Pre-training on Indoor Segmentation?},
  author    = {McCormac, John and Handa, Ankur and Leutenegger, Stefan and Davison, Andrew J.},
  booktitle = {ICCV},
  year      = {2017}
}

@inproceedings{zhong2025internscenes,
  title     = {{InternScenes}: A Large-scale Simulatable Indoor Scene Dataset with Realistic Layouts},
  author    = {Zhong, Weipeng and Cao, Peizhou and Jin, Yichen and Luo, Li and Cai, Wenzhe and Lin, Jingli and Wang, Hanqing and Lyu, Zhaoyang and Wang, Tai and Dai, Bo and Xu, Xudong and Pang, Jiangmiao},
  booktitle = {NeurIPS},
  year      = {2025}
}

@inproceedings{jiang2025megasynth,
  title     = {{MegaSynth}: Scaling Up {3D} Scene Reconstruction with Synthesized Data},
  author    = {Jiang, Hanwen and Xu, Zexiang and Xie, Desai and Chen, Ziwen and Jin, Haian and Luan, Fujun and Shu, Zhixin and Zhang, Kai and Bi, Sai and Sun, Xin and Gu, Jiuxiang and Huang, Qixing and Pavlakos, Georgios and Tan, Hao},
  booktitle = {CVPR},
  year      = {2025}
}

@inproceedings{xia2026sage,
  title     = {{SAGE}: Scalable Agentic {3D} Scene Generation for Embodied {AI}},
  author    = {Xia, Hongchi and Li, Xuan and Li, Zhaoshuo and Ma, Qianli and Xu, Jiashu and Liu, Ming-Yu and Cui, Yin and Lin, Tsung-Yi and Ma, Wei-Chiu and Wang, Shenlong and Song, Shuran and Wei, Fangyin},
  booktitle = {CVPR},
  year      = {2026}
}

@inproceedings{yang2024holodeck,
  title     = {{Holodeck}: Language Guided Generation of {3D} Embodied {AI} Environments},
  author    = {Yang, Yue and Sun, Fan-Yun and Weihs, Luca and VanderBilt, Eli and Herrasti, Alvaro and Han, Winson and Wu, Jiajun and Haber, Nick and Krishna, Ranjay and Liu, Lingjie and Callison-Burch, Chris and Yatskar, Mark and Kembhavi, Aniruddha and Clark, Christopher},
  booktitle = {CVPR},
  year      = {2024}
}

@inproceedings{yang2024physcene,
  title     = {{PhyScene}: Physically Interactable {3D} Scene Synthesis for Embodied {AI}},
  author    = {Yang, Yandan and Jia, Baoxiong and Zhi, Peiyuan and Huang, Siyuan},
  booktitle = {CVPR},
  year      = {2024}
}

@inproceedings{pfaff2026scenesmith,
  title     = {{S}cene{S}mith: Agentic Generation of Simulation-Ready Indoor Scenes},
  author    = {Pfaff, Nicholas and Cohn, Thomas and Zakharov, Sergey and Cory, Rick and Tedrake, Russ},
  booktitle = {ICML},
  year      = {2026}
}

@inproceedings{khanna2024hssd,
  title     = {{Habitat Synthetic Scenes Dataset} ({HSSD-200}): An Analysis of {3D} Scene Scale and Realism Tradeoffs for {ObjectGoal} Navigation},
  author    = {Khanna, Mukul and Mao, Yongsen and Jiang, Hanxiao and Haresh, Sanjay and Shacklett, Brennan and Batra, Dhruv and Clegg, Alexander and Undersander, Eric and Chang, Angel X. and Savva, Manolis},
  booktitle = {CVPR},
  year      = {2024}
}

@inproceedings{miao2026interiorgs,
  title     = {Towards Physically Executable {3D} {Gaussian} for Embodied Navigation},
  author    = {Miao, Bingchen and Wei, Rong and Ge, Zhiqi and Sun, Xiaoquan and Gao, Shiqi and Zhu, Jingzhe and Wang, Renhan and Tang, Siliang and Xiao, Jun and Tang, Rui and Li, Juncheng},
  booktitle = {ICLR},
  year      = {2026}
}

@inproceedings{wimbauer2026stepper,
  title     = {{Stepper}: Stepwise Immersive Scene Generation with Multiview Panoramas},
  author    = {Wimbauer, Felix and Manhardt, Fabian and Oechsle, Michael and Kalischek, Nikolai and Rupprecht, Christian and Cremers, Daniel and Tombari, Federico},
  booktitle = {CVPR Findings},
  year      = {2026}
}

@inproceedings{li2026argus,
  title     = {{Argus}: Metric Panoramic {3D} Reconstruction for Indoor Scenes},
  author    = {Li, Xi and Li, Linyuan and Wu, Yan and Rao, Tong and Zhang, Kai and Hui, Xinchen and Pan, Cihui},
  booktitle = {ECCV},
  year      = {2026}
}

@article{yu2026splatanything,
  title     = {{SplatAnything}: {Gaussian} Spacing Normalization for Feed-Forward {3DGS} across Arbitrary Scenes},
  author    = {Yu, Hao and Wang, Zhenwei and Hu, Yongzhen and Gong, Yuning and Wu, Junta and Zhou, Xiaowei and Bao, Hujun and Wang, Tengfei and Guo, Chunchao and Peng, Sida},
  journal   = {arXiv},
  year      = {2026}
}

@inproceedings{jin2025lvsm,
  title     = {{LVSM}: A Large View Synthesis Model with Minimal {3D} Inductive Bias},
  author    = {Jin, Haian and Jiang, Hanwen and Tan, Hao and Zhang, Kai and Bi, Sai and Zhang, Tianyuan and Luan, Fujun and Snavely, Noah and Xu, Zexiang},
  booktitle = {ICLR},
  year      = {2025}
}

@inproceedings{zhao2026erayzer,
  title     = {{E-RayZer}: Self-supervised {3D} Reconstruction as Spatial Visual Pre-training},
  author    = {Zhao, Qitao and Tan, Hao and Wang, Qianqian and Bi, Sai and Zhang, Kai and Sunkavalli, Kalyan and Tulsiani, Shubham and Jiang, Hanwen},
  booktitle = {CVPR},
  year      = {2026}
}

@inproceedings{paschalidou2021atiss,
  title     = {{ATISS}: Autoregressive Transformers for Indoor Scene Synthesis},
  author    = {Paschalidou, Despoina and Kar, Amlan and Shugrina, Maria and Kreis, Karsten and Geiger, Andreas and Fidler, Sanja},
  booktitle = {NeurIPS},
  year      = {2021}
}

@inproceedings{zhang2017pbrs,
  title     = {Physically-Based Rendering for Indoor Scene Understanding Using Convolutional Neural Networks},
  author    = {Zhang, Yinda and Song, Shuran and Yumer, Ersin and Savva, Manolis and Lee, Joon-Young and Jin, Hailin and Funkhouser, Thomas},
  booktitle = {CVPR},
  year      = {2017}
}

@inproceedings{li2021openrooms,
  title     = {{OpenRooms}: An Open Framework for Photorealistic Indoor Scene Datasets},
  author    = {Li, Zhengqin and Yu, Ting-Wei and Sang, Shen and Wang, Sarah and Song, Meng and Liu, Yuhan and Yeh, Yu-Ying and Zhu, Rui and Gundavarapu, Nitesh and Shi, Jia and Bi, Sai and Yu, Hong-Xing and Xu, Zexiang and Sunkavalli, Kalyan and Ha{\v{s}}an, Milo{\v{s}} and Ramamoorthi, Ravi and Chandraker, Manmohan},
  booktitle = {CVPR},
  year      = {2021}
}

@inproceedings{zhu2022learning,
  title     = {Learning-Based Inverse Rendering of Complex Indoor Scenes with Differentiable {Monte Carlo} Raytracing},
  author    = {Zhu, Jingsen and Luan, Fujun and Huo, Yuchi and Lin, Zihao and Zhong, Zhihua and Xi, Dianbing and Zheng, Jiaxiang and Tang, Rui and Bao, Hujun and Wang, Rui},
  booktitle = {SIGGRAPH Asia},
  year      = {2022}
}

@inproceedings{ou2026holo360d,
  title = {{Holo360D}: A Large-Scale Real-World Dataset with Continuous Trajectories for Advancing Panoramic {3D} Reconstruction},
  author = {Ou, Jing and Cao, Zidong and Ren, Yinrui and Li, Zhuoxiao and Zhu, Jinjing and Hua, Tongyan and Zhang, Shuai and Xiong, Hui and Zhao, Wufan},
  booktitle = {ECCV},
  year = {2026}
}

@article{yang2025matrix3d,
  title     = {{Matrix-3D}: Omnidirectional Explorable {3D} World Generation},
  author    = {Yang, Zhongqi and Ge, Wenhang and Li, Yuqi and Chen, Jiaqi and Li, Haoyuan and An, Mengyin and Kang, Fei and Xue, Hua and Xu, Baixin and Yin, Yuyang and Li, Eric and Liu, Yang and Wang, Yikai and Guo, Hao-Xiang and Zhou, Yahui},
  journal   = {arXiv},
  year      = {2025}
}

@inproceedings{huang2026omnix,
  title     = {{OmniX}: From Unified Panoramic Generation and Perception to Graphics-Ready {3D} Scenes},
  author    = {Huang, Yukun and Yu, Jiwen and Zhou, Yanning and Wang, Jianan and Wang, Xintao and Wan, Pengfei and Liu, Xihui},
  booktitle = {ECCV},
  year      = {2026}
}

@inproceedings{liu2026cmevs,
  title     = {{CM-EVS}: Sparse Panoramic {RGB-D-Pose} Data for Coverage-Oriented Scene Observation},
  author    = {Liu, Jiale and Li, Jungang and Yu, Jieming and Yu, Xinglin and Dongfang, Zihao and Shi, Keyu and Ding, Zongjian and Zhang, Jiahuan and Bai, Shunwen and Huang, Haoran and Wang, Yurun and Wu, Yanxi and Yu, Ningzhe and Gao, Yudong and Cheng, Mingjun},
  booktitle = {NeurIPS},
  year      = {2026}
}

@inproceedings{bozic2026pager,
  title     = {Unified Panoramic Geometry Estimation via Multi-View Foundation Models},
  author    = {Bozic, Vukasin and Slavkovic, Isidora and Narnhofer, Dominik and Metzger, Nando and Rozumny, Denis and Schindler, Konrad and Kalischek, Nikolai},
  booktitle = {NeurIPS},
  year      = {2026}
}
